\documentclass[10pt,twocolumn,letterpaper]{article}

\usepackage[pagenumbers]{wacv} 

\usepackage{graphicx}
\usepackage{amsmath}
\usepackage{amssymb}
\usepackage{booktabs}
\usepackage{url}
\usepackage{colortbl}
\usepackage{multirow}
\usepackage[table]{xcolor}
\usepackage[pagebackref,breaklinks,colorlinks]{hyperref}

\usepackage[capitalize]{cleveref}
\crefname{section}{Sec.}{Secs.}
\Crefname{section}{Section}{Sections}
\Crefname{table}{Table}{Tables}
\crefname{table}{Tab.}{Tabs.}

\def\wacvPaperID{1916} 
\def\confName{WACV}
\def\confYear{2024}

\begin{document}

\title{Bridging the Gap between Multi-focus and Multi-modal: A Focused Integration Framework for Multi-modal Image Fusion}

\author{
Xilai Li\textsuperscript{1}\hspace{1em}
Xiaosong Li\textsuperscript{1}\thanks{Corresponding author.}\hspace{1em}
Tao Ye\textsuperscript{2}\hspace{1em}
Xiaoqi Cheng\textsuperscript{1}\hspace{1em}
Wuyang Liu\textsuperscript{1}\hspace{1em}
Haishu Tan\textsuperscript{1}\\
{\small\textsuperscript{1} Foshan University, Foshan 528225, China}\\
{\small\textsuperscript{2} China University of Mining and Technology, Beijing 100083, China}\\
{\tt\small 20210300236@stu.fosu.edu.cn, lixiaosong@buaa.edu.cn}
}
\maketitle
\begin{abstract}
Multi-modal image fusion (MMIF) integrates valuable information from different modality images into a fused one. However, the fusion of multiple visible images with different focal regions and infrared images is a unprecedented challenge in real MMIF applications. This is because of the limited depth of the focus of visible optical lenses, which impedes the simultaneous capture of the focal information within the same scene. To address this issue, in this paper, we propose a MMIF framework for joint focused integration and modalities information extraction. Specifically, a semi-sparsity-based smoothing filter is introduced to decompose the images into structure and texture components. Subsequently, a novel multi-scale operator is proposed to fuse the texture components, capable of detecting significant information by considering the pixel focus attributes and relevant data from various modal images. Additionally, to achieve an effective capture of scene luminance and reasonable contrast maintenance, we consider the distribution of energy information in the structural components in terms of multi-directional frequency variance and information entropy. Extensive experiments on existing MMIF datasets, as well as the object detection and depth estimation tasks, consistently demonstrate that the proposed algorithm can surpass the state-of-the-art methods in visual perception and quantitative evaluation. The code is available at \textcolor{red}{\href{https://github.com/ixilai/MFIF-MMIF}{https://github.com/ixilai/MFIF-MMIF}}.
\end{abstract}

\section{Introduction}
\label{sec:intro}
Multi-modal image fusion (MMIF) techniques play a crucial role in computer vision by integrating valuable information from diverse sensors. These techniques provide a comprehensive and detailed interpretation of scenes for advanced vision tasks such as target detection \cite{C1,c59,c60}, semantic segmentation \cite{C2,c57,c58}, and pedestrian re-identification \cite{C3}. Take the infrared and visible image fusion (IVIF) task for example, visible sensors perform well in well-lit scenes, capturing rich textures and details. However, they lose information in low light or harsh conditions such as smoke or rain. In contrast, thermal imaging cameras are less susceptible to outside interference and can capture thermal radiation from a target, but they lack sensitivity to complex scene details.

\begin{figure}[t]
  \centering
   \includegraphics[width=1.0\linewidth]{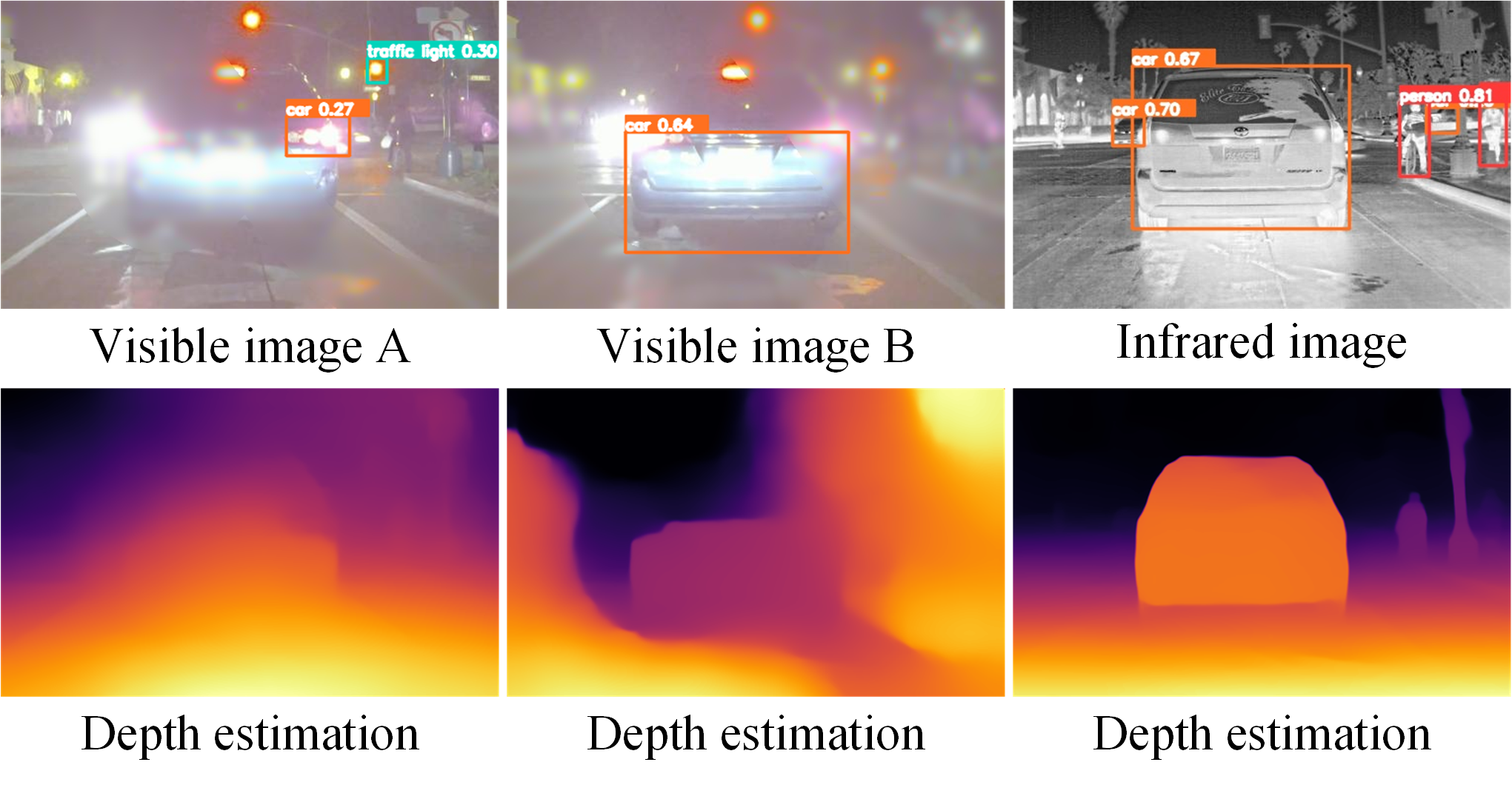}
   \caption{Application of multi-modal images to object detection and depth estimation tasks.}
   \label{fig:onecol0}
\end{figure}

In recent years, a number of fusion algorithms have been developed in the field of IVIF. They can be broadly classified into deep learning (DL)-based methods \cite{C4,C5,C43} and general fusion methods \cite{C6,C7,C8}. DL-based methods employ neural networks to emulate human brain functions. By learning from extensive datasets, these models establish connections and utilize deep features to reconstruct fused images rich in intricate details. Currently, various network modules initially designed for low-level vision tasks have been successfully adapted to IVIF tasks. These modules include Attention Mechanisms \cite{C24, C44, C45}, Dilated Convolutions \cite{C46}, Encoder-Decoder Models \cite{C47, C48, C49}, and Generative Models \cite{C50, C51, C52}. General fusion algorithms can be divided into multi-scale transform (MST) \cite{C10, C54} and saliency-based \cite{C11, C53} algorithms. The MST algorithm decomposes the source image into multiple scales and directions, integrates them using fusion rules, and reconstructs the fusion coefficients. The better scale-awareness of such methods is also accompanied by higher computational costs. The saliency-based algorithm preserves salient regions from different source images using a saliency map and fusion weights, reducing pixel redundancy and improving fusion result visual quality.

Although the algorithms \cite{C4,C9,C10,C11,C12} mentioned above can deliver high-quality fusion results, they usually assume that the scene information captured by the visible imaging device is always in focus and sharp. However, in real situations, owing to the limitations of optical lenses, only objects within the depth of field can be considered. As shown in \cref{fig:onecol0}, the challenge arises when the visible image's focus area is incomplete, leading to difficulties in accurately providing target and depth information for advanced tasks.  Therefore, when the camera fails to capture all the target information in the scene simultaneously, multiple sets of data must be captured to ensure that all the target information is within the focus area. To address these challenges, we propose a focusing information integration framework for the MMIF. This framework aims to simultaneously integrate clear pixel information from different focusing regions and salient pixel information captured from different modalities. Our method decomposes the source image into texture and structure components. For texture fusion, we employ a salient feature extraction operator guided by Gaussian and Laplacian pyramid levels, capturing salient details across scales. We also utilize multi-directional frequency variance and information entropy to identify high-energy pixel information in the structure components. The contributions of this study can be summarized as follows:

(1) We propose a focused integration framework for MMIF. To the best of our knowledge, this is the first attempt to address the multi-focus interference for MMIF task, accurately achieve simultaneous synthesis of focused and multi-modal information.

(2) We propose a feature extraction operator based on pyramid scale separation, which fully considers both pixel focus attributes and saliency to effectively capture the distribution of details in each source image.

(3) We generate a dataset which integrates the multi-focus image fusion and MMIF tasks, effectively reflected the defocus fusion challenge in MMIF realworld applications and provided a new benchmark for evaluating and improving existing algorithms.

(4) Experimental results yield highly competitive performance on multiple computer vision tasks such as MMIF, as well as object detection and depth estimation, with both subjective and objective evaluations demonstrating the effectiveness of the proposed method.

\begin{figure*}[h]
  \centering
   \includegraphics[width=1.0\linewidth]{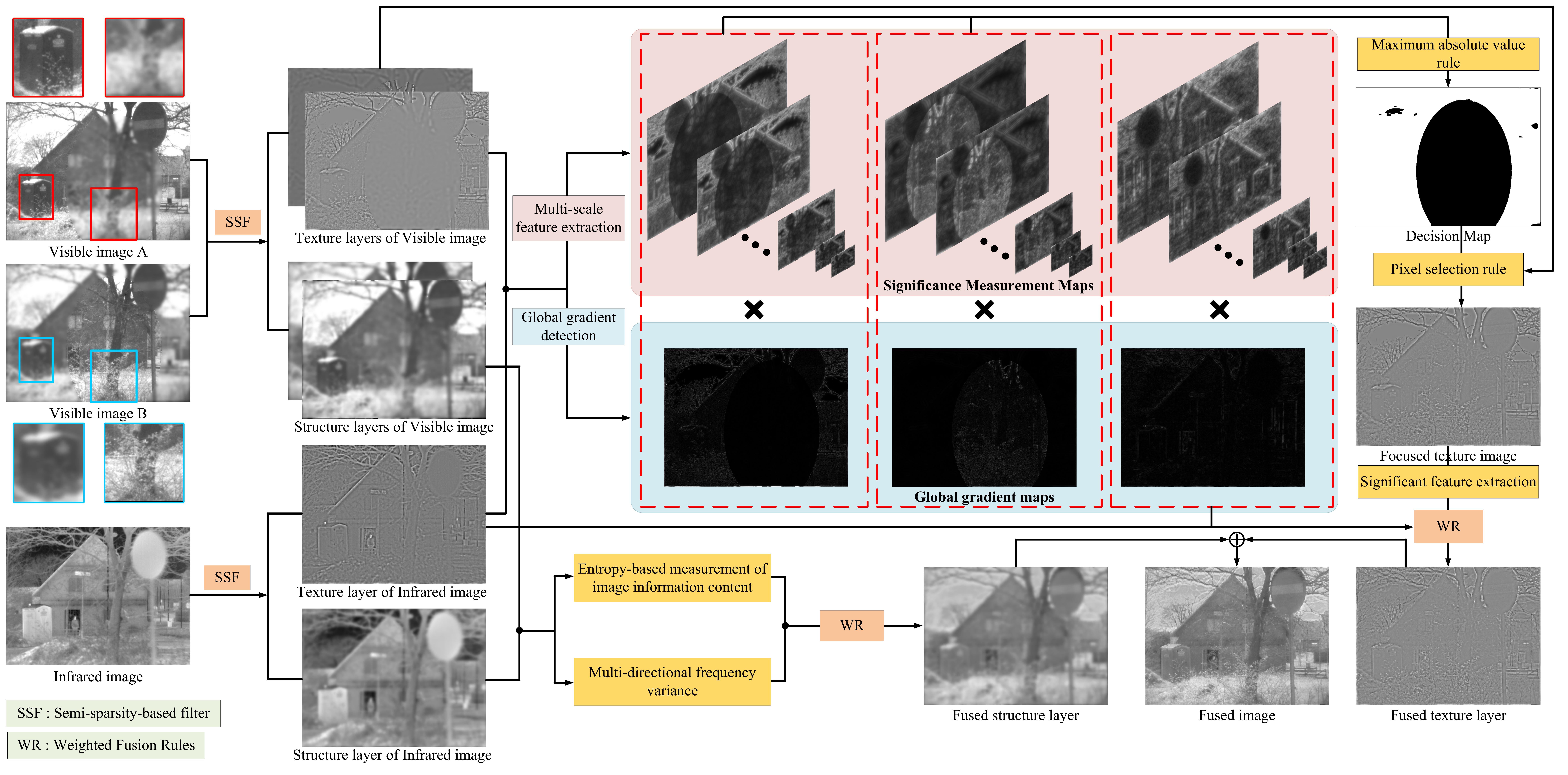}
   \caption{The flowchart of the proposed method.}
   \label{fig:onecol1}
\end{figure*}

\section{Related Work}
\subsection{DL-based methods}
In the field of IVIF, deep learning has emerged as a powerful tool, harnessing its strong representation learning capabilities and high adaptability \cite{C55, C20, C23, C24, C25}. For example, Tang et al. \cite{C9} proposed a semantic-aware real-time IVIF framework that leveraged a segmentation network to predict the segmentation result of a fused image, allowing semantic loss to be constructed. This loss is then utilized to guide the training of the fusion network through backpropagation. Ma et al. \cite{C4} presented a Darkness-free IVIF algorithm that ad-dresses the issue of degraded illumination features in source images. Their approach involved the design of a scene illumination network that effectively removed the degraded features. Furthermore, the algorithm focused on enhancing the visual quality of the fusion results by leveraging the unique characteristics of different modal images. Zhao et al.\cite{C47}proposed an IVIF algorithm combining convolutional neural network and transformer, which can effectively model cross-modal images and distinguish between public and private features between different modal images. Liu et al. \cite{C50} tackled the challenges of IVIF in target detection, and proposed a target-aware dual adversarial learning fusion network and a commonly used detection network that utilizes a generator and dual discriminators to learn commonalities between different modal images in order to effectively retain salient pixel information within the scene.
\subsection{Semi-sparse filter}
Recent studies \cite{C13} show that sparse priors in edge-preserving filters can effectively achieve segmented constant filtering outcomes, preserving sparse features like singularities and edges. However, the sparsity assumption may fail in regions with sparse features and polynomial-smoothing surfaces. Based on the above, Huang et al \cite{C14} demonstrated that a smoothing function constructed using a semi-sparse prior can be applied to the case of sparsity prior failure, and this semi-sparse minimization scheme is expressed as follows:
\begin{equation}
  \min_u \left\|u-Zf\right\|_2^2 + \alpha\sum_{k=1}^{n-1}\left\|\nabla^k u-\nabla^k (Zf)\right\|_2^2 + \lambda\left\|\nabla^n u\right\|_0
  \label{eq:important}
\end{equation}

\noindent where $f$ and $u$ denote the input and output images, respectively, the weights $Z$ \cite{C14} specify the confidence map of the spatial variation of the smoothness, $\alpha$ and $\lambda$ are the balanced weights, which are set to 0.8 and 0.05, respectively. $\nabla^n$ represents the $n$-th (partial) differential operator. In the proposed algorithm, we set $n$=2 as the highest order of regularization. Finally, we can express the semi-sparsity filter (SSF) as follows,
\begin{equation}
  u = SSF(f)
  \label{eq:important}
\end{equation}

\noindent where $\text{\textit{SSF}}(\cdot)$ represents the semi-sparsity filtering operator; detailed information about this filter can be found in \cite{C14}.

\section{Proposed Method}
\cref{fig:onecol1} shows a flowchart of the proposed algorithm; it can input two visible images with different focus regions as well as a corresponding infrared image and then output a fused image with a clear full scene and rich feature information.

\subsection{Image decomposition}

The edge-preserving filter can effectively smooth most of the texture and structure information in the source image, while maintaining the intensity of the structured edges. To better distinguish the pixel points representing different feature information in the image, we first decomposed the source image using SSF. The visible images $f_1$ and $f_2$ in different focus regions and the corresponding infrared images $f_3$ are input, the operation for acquiring structural layer $S$ is as follows:
\begin{equation}
  S_m = SSF(f_m)
  \label{eq:important}
\end{equation}

\noindent where $m \in \{1, 2, 3\}$, $S_m$ represents the structural layers corresponding to different source images, and the texture layers are computed as follows,
\begin{equation}
  T_m = f_m-S_m
  \label{eq:important}
\end{equation}

\noindent where $T_m$ denotes the texture layers.

\subsection{Significant feature fusion}

\subsubsection{Texture layer fusion}

The fusion of texture information relies on detecting focused and sharp detail information as well as on leveraging complementary information between different modal images. To address this challenge, we designed a novel feature extraction operator comprising of two key components, global gradient information detection and multiscale feature extraction. These components work together to effectively extract and integrate texture details from input images. First, we obtain the gradient information of the texture layer using a global gradient detection operator, which is represented as follows:
\begin{equation}
  GM_{*,m} = r \nabla T_{*,m} - (2 \nabla T_{*,m} + 0.01)^p
  \label{eq:important}
\end{equation}

\noindent where $\nabla T_*$ ($* \in \{x, y\}$) represents the gradient of $T$ along the $x$-axis or $y$-axis, $p$ is a constant that is set to 0.8, and $r = p \times e^{\frac{p}{2}-1}$. We utilized the discrete differential operators $[-1,1]$ and $[1,-1]^T$ to calculate the gradients along the $x$-axis and $y$-axes, respectively. Guided by the gradient information, we can effectively extract the clear details of the focus.

To detect complementary information across different modal images, we propose a scale feature extraction operator based on Gaussian and Laplacian pyramids. The Gaussian pyramid generates a multi-scale representation by consistently downsampling the image. Contrastingly, the Laplacian pyramid is derived from the Gaussian pyramid and represents the residual between the Gaussian pyramid image and the upsampled image from the previous level. By exploiting the inherent scale-separation property of a Gaussian pyramid, we can effectively extract feature information at various scales. Given an input image $T$, it is decomposed into $N$ scales $G_l$, $l \in \{0, 1, ..., N\}$ by using a Gaussian pyramid, where $G_0 = T$. Each level of the Laplace pyramid corresponds to $L_l$, $l \in \{0, 1, ..., N-1\}$, and our goal is to detect the significant pixel information of the image guided by the scale $k$ ($0 < k < N$). The detailed process is shown below:
\begin{equation}
  SM_m = \sum_{k=N-1}^{1} \sqrt{SF(G_{m,k}) + SF(L_{m,k})}
  \label{eq:important}
\end{equation}

\noindent where $G_{m,k}$ and $L_{m,k}$ represent the $k$-th Gaussian and Laplace pyramids corresponding to the texture layer $T_m$, respectively. $SF(\cdot)$ represents the spatial frequency operator \cite{C15} used to detect significant feature information and $SM$ represents the significance measure map. Additionally, $G_{m,k}$ and $L_{m,k}$ are upsampled after each feature extraction to recover the same size as the previous layer. Finally, by combining the global gradient and significance measures, the novel feature extraction operator proposed in this study can be expressed as follows:
\begin{equation}
  TM_m = SM_m \times \sqrt{GM_{x,m}^2 + GM_{y,m}^2}
  \label{eq:important}
\end{equation}

\noindent where $TM_m$ denotes the salient feature map. Under the guidance of $TM_m$, we first obtain the focus decision map using the “maximum absolute value rule”.  
\begin{equation}
 \text{$MAP$}(x, y) = \begin{cases}
1, & \text{\textit{if }} TM_1(x, y) > TM_2(x, y) \\
0, & \text{otherwise}
\end{cases}
  \label{eq:important}
\end{equation}

\noindent we then introduced a multiscale consistency verification technique \cite{C16} to process the $MAP$ and obtain $OMP$, according to which we can integrate the focused details and obtain the focused texture map $T_4$.
\begin{equation}
  \text{$T_4$}(x, y) = \begin{cases}
T_1(x, y), & \text{\textit{if }} \text{$MAP$}(x, y) = 1 \\
T_2(x, y), & \text{otherwise}
\end{cases}
  \label{eq:important}
\end{equation}

\noindent After obtaining the salient feature map $TM_4$ of $T_4$ using the aforementioned feature extraction operator, the final fused texture layer $FT$ is constructed using the following rules:
\begin{equation}
  FT = \sum_{m=3}^4 \left( \frac{TM_m}{TM_3 + TM_4} \times T_m \right)
  \label{eq:important}
\end{equation}

\subsubsection{Structure layer fusion}

The structural layer contained low-frequency information from the source image. Inspired by \cite{C17}, our study considers the distribution of energy information in terms of both the entropy and multidirectional frequency variance of the image. Firstly, we calculate the frequency variance $\psi_m$ for each $3 \times 3$ sized discrete cosine transform (DCT) \cite{C18} block in the structural layer.
\begin{equation}
  \psi_m = \frac{\sigma_m^\theta}{\varphi_m}
  \label{eq:important}
\end{equation}

\noindent where $\sigma_m^\theta$ represents the standard deviation of the DCT blocks in four directions $\theta \in \{0^\circ, 45^\circ, 90^\circ, 135^\circ\}$, and $\varphi_m$ represents the mean of $\sigma_m^\theta$ in four directions. We calculated the variance of all the blocks and then computed their average values as the first eigenvalue of the structural layer.

The entropy of an image is used to measure its information richness. To enable the proposed algorithm to handle more complex images efficiently, we used entropy as a second feature value for designing the structural layer fusion rules. Therefore, the fused structural layer $FS$ is constructed as follows.
\begin{equation}
FS = \sum_{m=1}^3 \left( \frac{{E_m \times \psi_m}}{{K }}\times S_m \right)
  \label{eq:important}
\end{equation}

\noindent where $K = \sum_{m=1}^3 (E_m \times \psi_m)$, $E_m$ represents the entropy of $S_m$.

\subsection{Fusion result construction}
Based on $FT$ and $FS$, the final fusion result $F$ can be constructed.
\begin{equation}
F = FT + FS
  \label{eq:important}
\end{equation}
For the proposed model, the fusion process of Visible Image A and Infrared Image can be used for the MMIF task in the normal case, and the fusion process of Visible Image A and Visible Image B can be used for the task of multi-focus image fusion.

\begin{figure}[h]
  \centering
   \includegraphics[width=1.0\linewidth]{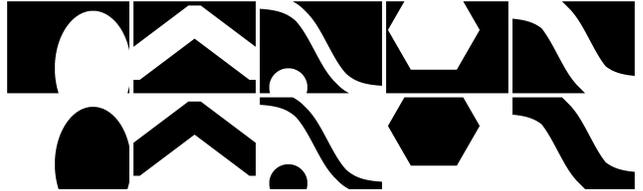}
   \caption{Examples of binary mask pairs.}
   \label{fig:onecol2}
\end{figure}

\section{Dataset generation}
We generated a dataset by integrating the MFIF and MMIF tasks based on two publicly available MMIF datasets, TNO \cite{C19} and RoadScene \cite{C20}. Inspired by \cite{C21}, we first randomly generated a pair of binary masks based on the size of the source image and subsequently used them to blur two complementary regions in the visible images to simulate a realistic situation in which the focused region of the scene is incomplete. The specific implementation is as follows.
\begin{equation}
f_1 = f_{\text{clear}} \times M_1 + f_{\text{blur}} \times M_2
  \label{eq:important}
\end{equation}
\begin{equation}
f_2 = f_{\text{clear}} \times M_2 + f_{\text{blur}} \times M_1
  \label{eq:important}
\end{equation}

\noindent where $M_1$ and $M_2$ represent a pair of complementary masks and the blurred image $f_{blur}$ is obtained by convolving the fully focused visible image $f_{clear}$ with a Gaussian filter kernel with a standard deviation of 5. Thus, we can decompose a fully focused visible image into a set of multi-focused source images with complementary focus regions. \cref{fig:onecol2} shows a partially randomly generated binary mask.

\section{Experiment}
\subsection{Experimental Setting}
We used seven state-of-the-art MMIF image fusion methods for comparison, which are TarD \cite{C50}, ReC \cite{C46}, CDD \cite{C47}, LRR \cite{C43}, SeA \cite{C9}, SDD \cite{C25}, and U2F \cite{C20}. All methods were tested on the source code provided by the original authors, with a PC with a NVIDIA GeForce RTX 3060Ti GPU and a Gen Intel Core i7-13700F CPU. Additionally, we selected six objective evaluation metrics to assess the fusion performance of different algorithms, which are, $Q_G$ \cite{C26}, $Q_M$ \cite{C27}, $Q_S$ \cite{C28}, $AG$ \cite{C29}, $SF$ \cite{C15}, and $PSNR$ \cite{C30}. These six metrics can be used to comprehensively evaluate the quality of different fusion results from several aspects, with higher metric scores representing better image quality.

Moreover, to ensure experimental fairness, except for the proposed algorithm, whose input images are the multi-focus images and the corresponding infrared images, the inputs of the remainder of the comparison methods are only two images, the fully focused visible images and the corresponding infrared images. In addition, all the images used in this study are derived from both the TNO and RoadScene datasets.

\begin{figure}[h]
  \centering
   \includegraphics[width=1.0\linewidth]{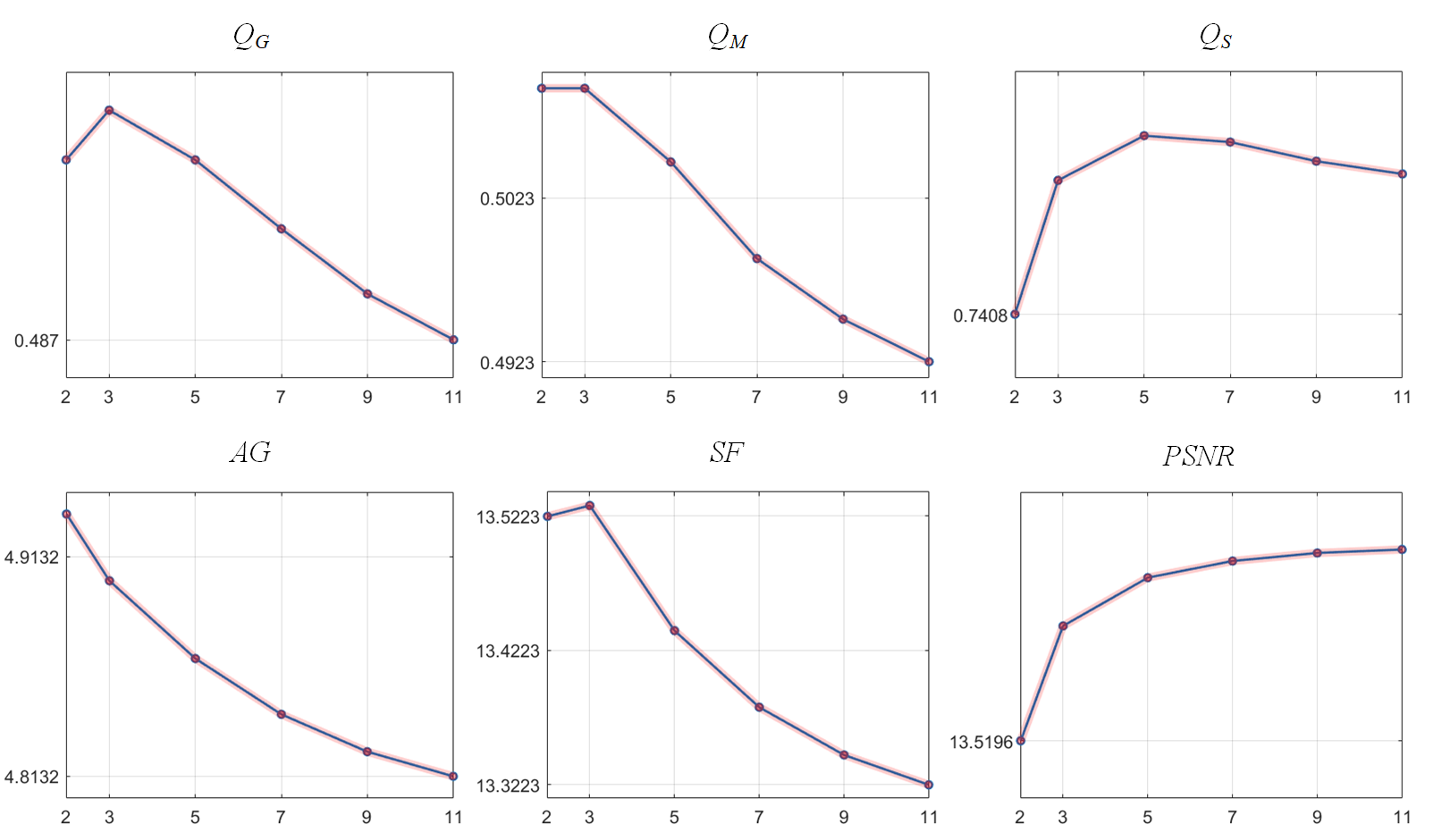}
   \caption{The quantitative comparison results of the proposed algorithm for different values of the parameter $N$.}
   \label{fig:onecol3}
\end{figure}

\subsection{Parameter Analysis}
We conducted an analysis focusing on the number of pyramid decomposition layers, denoted as $N$. For this analysis, we randomly selected $20$ image pairs from the RoadScene dataset as a quantitative comparison dataset. The fusion performance under different $N$ values was evaluated using various metrics, and the corresponding scores are shown in \cref{fig:onecol3}. The horizontal axis represents the $N$ values, while the vertical axis represents the scores for each metric. From the results shown in \cref{fig:onecol3}, it is evident that three metrics achieved highest scores when $N$ was set to three. Based on this observation, we decided to set $N$ to three in our algorithm.

\begin{figure}[h]
  \centering
   \includegraphics[width=1.0\linewidth]{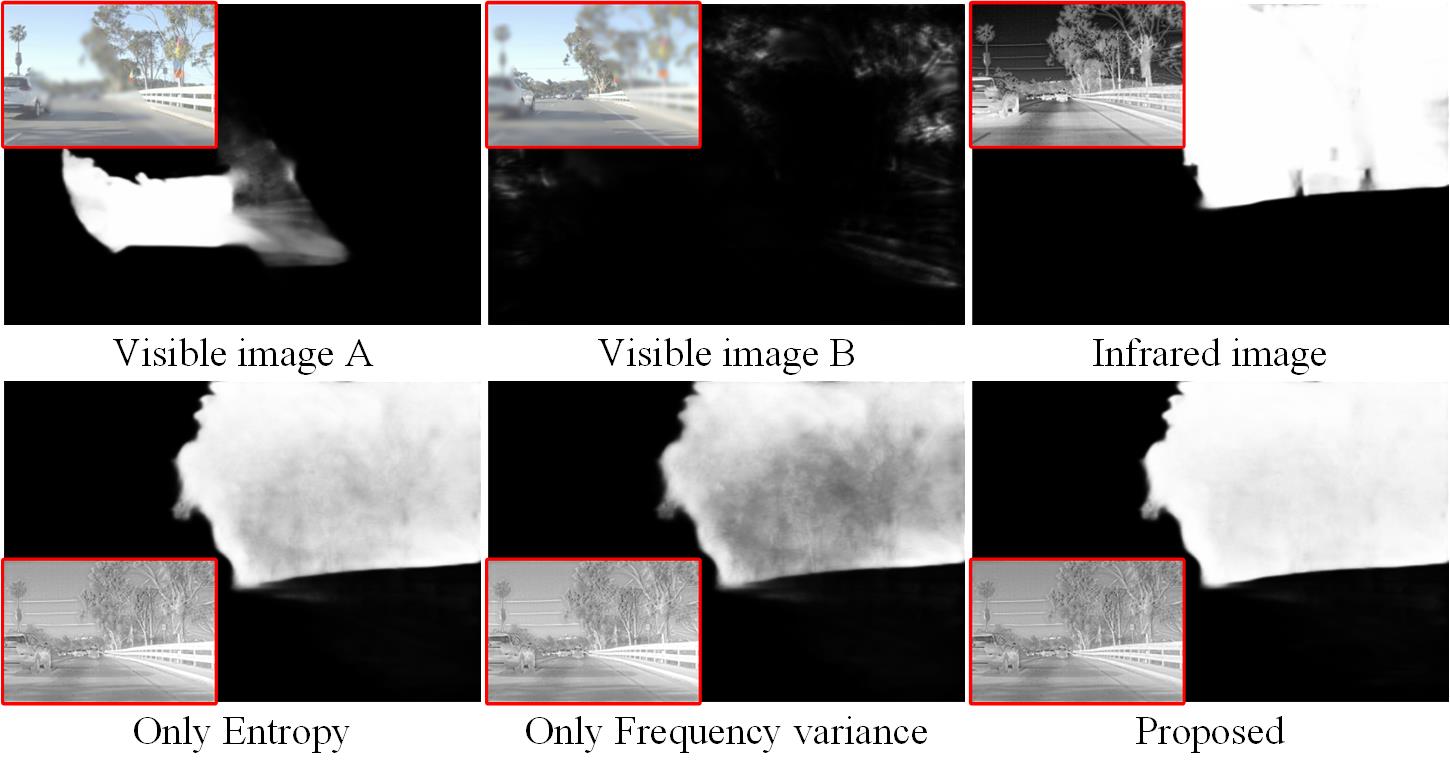}
   \caption{Ablation study of fusion rules for structural layers. The figure shows the salient target detection results corresponding to different source images and fusion results.}
   \label{fig:onecol10}
\end{figure}

\begin{figure*}[h]
  \centering
   \includegraphics[width=1.0\linewidth]{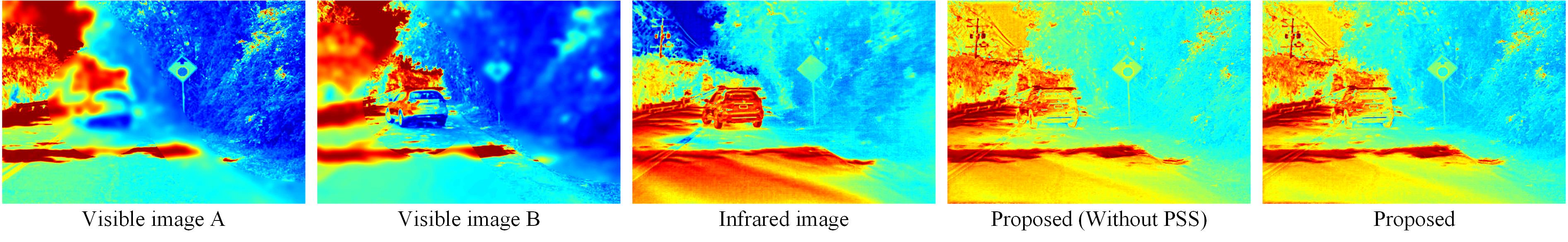}
   \caption{Ablation analysis of the pyramid scale separation. PSS: Pyramid Scale Separation.}
   \label{fig:onecol7}
\end{figure*}

\subsection{Ablation Analysis}
During the structural layer fusion process, we carefully consider the distribution of energy information from two perspectives: entropy and frequency variance. We conduct ablation experiments to validate the effectiveness of this strategy. We employ a salient target detection algorithm \cite{C56} to segment the various fusion results, evaluating their performance in extracting salient pixel information, as shown in the \cref{fig:onecol10}, the fusion rule designed by considering both perspectives demonstrates the best performance. 

In addition, we introduce a novel feature extraction algorithm based on pyramid scale separation. Ablation studies are conducted to assess the effectiveness of this strategy. \cref{fig:onecol7} presents a series of experimental results on the RoadScene dataset. Pseudo colors are assigned to both the source images and the fusion results for easy comparison. \cref{fig:onecol7} clearly demonstrates that the fusion results without pyramid scale separation lack certain detail information and fail to highlight the texture and hierarchical details of certain objects, such as leaves. Conversely, our proposed algorithm, leveraging multi-scale representation through pyramids, successfully captures subtle texture information. This observation confirms that our strategy significantly improves the detail-aware performance of the proposed algorithm.

\subsection{Comparison results on the TNO dataset}
\begin{figure*}[h]
  \centering
   \includegraphics[width=1.0\linewidth]{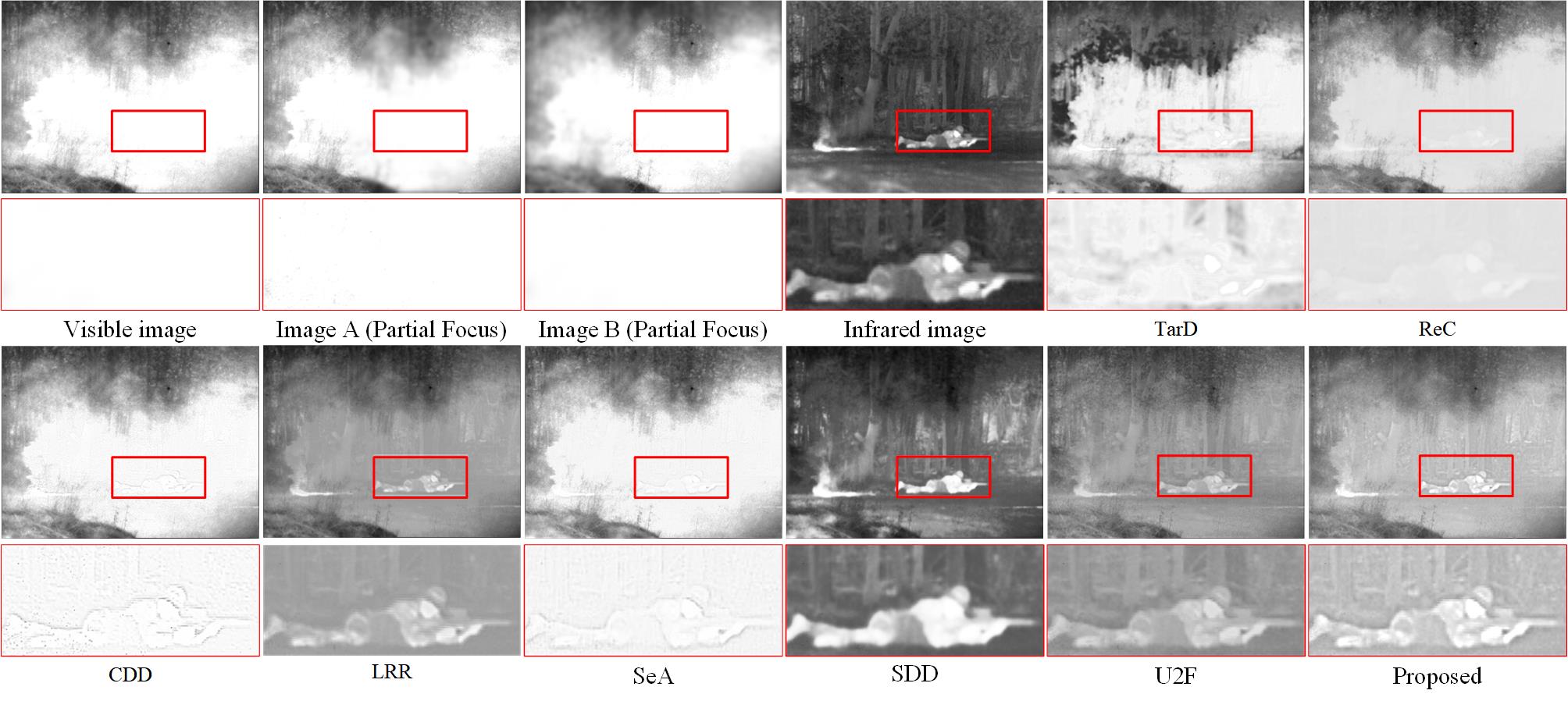}
   \caption{Qualitative comparison results of different algorithms on a set of typical image pairs in the TNO dataset.}
   \label{fig:onecol4}
\end{figure*}

\cref{fig:onecol4} shows the fusion results of the different algorithms for a typical scene and zooms in on the local area to better compare the useful information retained by the different algorithms. The challenge for each algorithm in this scene was to capture soldiers hidden in smoke while minimizing smoke interference in the visible image. In the highlighted enlarged area, TarD, ReC, CDD and SeA methods are severely interfered by smoke, resulting in poor contrast and inability to extract target information from the smoke. The LRR and SDD methods rely primarily on infrared image pixel information, resulting in the loss of visible image details. In contrast, our proposed algorithm effectively extracts useful pixel in-formation from different modal images without smoke interference and exhibits superior detail retention.

\cref{tab:comparison1} displays the quantitative comparison results of the proposed algorithm and seven other methods on the TNO dataset. The proposed algorithm outperformed the other algorithms in four metrics, demonstrating its capability to accurately detect clear pixels in diverse focus regions and effectively integrate information from different modalities. It surpassed current state-of-the-art fusion algorithms in preserving both detail and energy information.

\begin{table}[h]
\centering
\caption{Average scores by quantitative comparison on the TNO dataset. \textbf{Bold} is the best and \textcolor{red}{Red} is the second-best.}
\footnotesize 
\label{tab:comparison1}
\begin{tabular}{@{}l@{\hspace{1em}}llllll@{}}
\hline
Methods & $Q_G$ & $Q_M$ & $Q_S$ & $AG$ & $SF$ & $PSNR$ \\
\hline
TarD\cite{C50}(\textit{CVPR}) & 0.34 & 0.46 & 0.77 & 2.74 & 6.95 & 18.35 \\
ReC\cite{C46}(\textit{ECCV}) & 0.35 & 0.48 & \textcolor{red}{0.78} & 3.00 & 6.22 & 18.50 \\
CDD\cite{C47}(\textit{CVPR}) & 0.40 & 0.50 & 0.73 & \textcolor{red}{5.47} & \textbf{14.21} & 15.30 \\
LRR\cite{C43}(\textit{TPAMI}) & 0.33 & 0.41 & 0.68 & 4.34 & 10.43 & 14.68 \\
SeA\cite{C9}(\textit{Inf}) & 0.31 & 0.37 & 0.69 & \textbf{5.84} & \textcolor{red}{13.88} & 16.18 \\
SDD\cite{C25}(\textit{TMM}) & 0.31 & 0.46 & 0.70 & 4.26 & 9.99 & 14.23 \\
U2F\cite{C20}(\textit{TPAMI}) & \textcolor{red}{0.42} & \textcolor{red}{0.54} & \textbf{0.83} & 3.44 & 8.01 & \textcolor{red}{19.21} \\
\rowcolor{gray!20} Proposed & \textbf{0.43} & \textbf{0.64} & \textbf{0.83} & 3.39 & 9.40 & \textbf{19.39} \\
\hline
\end{tabular}
\end{table}

\subsection{Comparison results on the RoadScene dataset}
\begin{figure*}[h]
  \centering
   \includegraphics[width=1.0\linewidth]{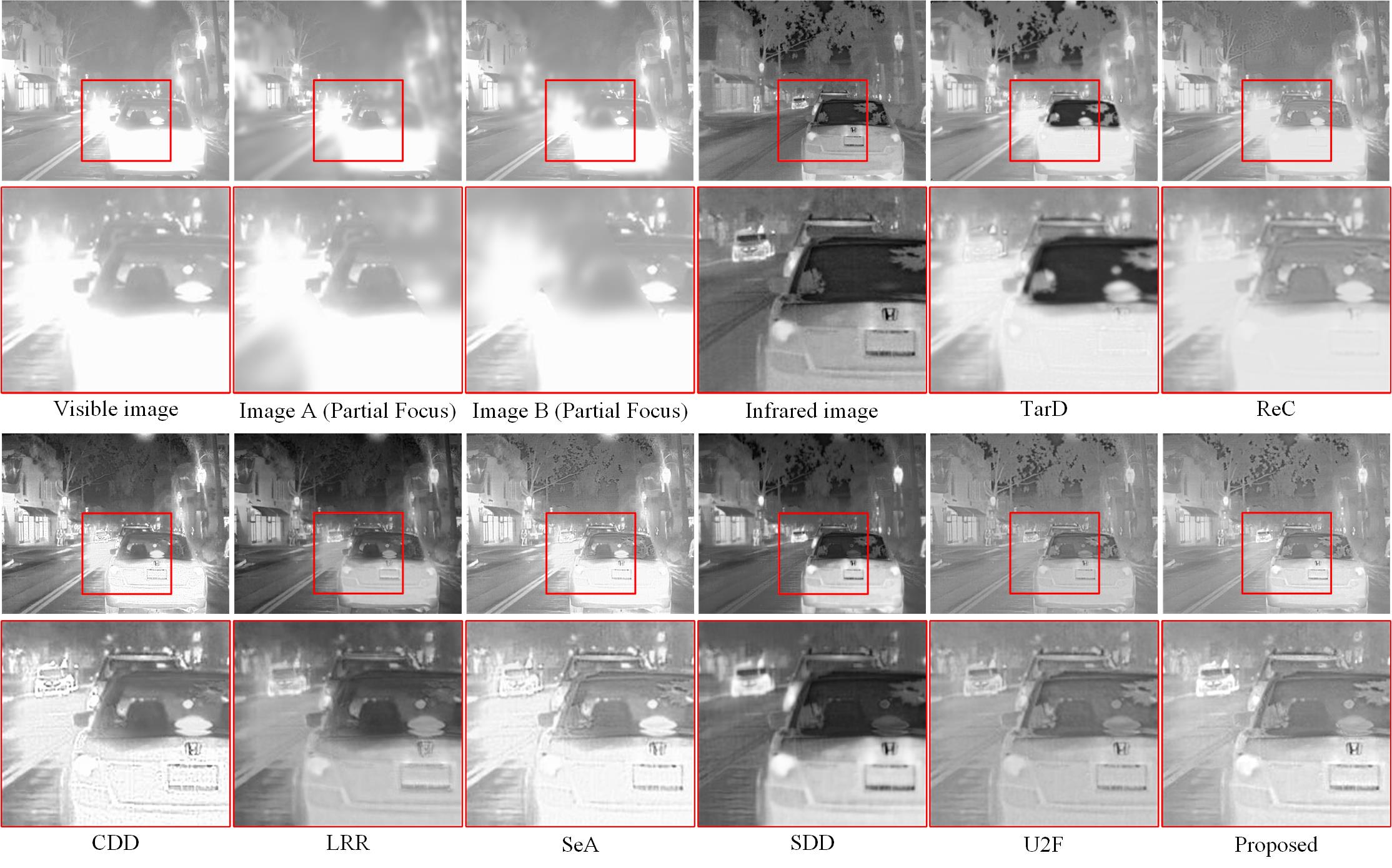}
   \caption{Qualitative comparison results of different algorithms on a set of typical image pairs in the RoadScene dataset.}
   \label{fig:onecol5}
\end{figure*}

\cref{fig:onecol5} illustrates the fusion results of various fusion algorithms on a set of image pairs from the road scene dataset. In this set, the visible images exhibit overexposure. As shown in the red enlarged area, the fusion results of the four methods, TarD, ReC, CDD and SeA ingest too much pixel information from the overexposed regions, which hinders the fusion results from maintaining a reasonable contrast. Both LRR and U2F methods were unable to recognize the weak texture information provided by the IR images, resulting in incomplete scene information. Conversely, the proposed algorithm demonstrated robustness in complex scenes, effectively extracting the salient pixel information that is crucial for advanced vision tasks, even in the presence of scene disturbances.

\begin{figure*}[t]
  \centering
   \includegraphics[width=1.0\linewidth]{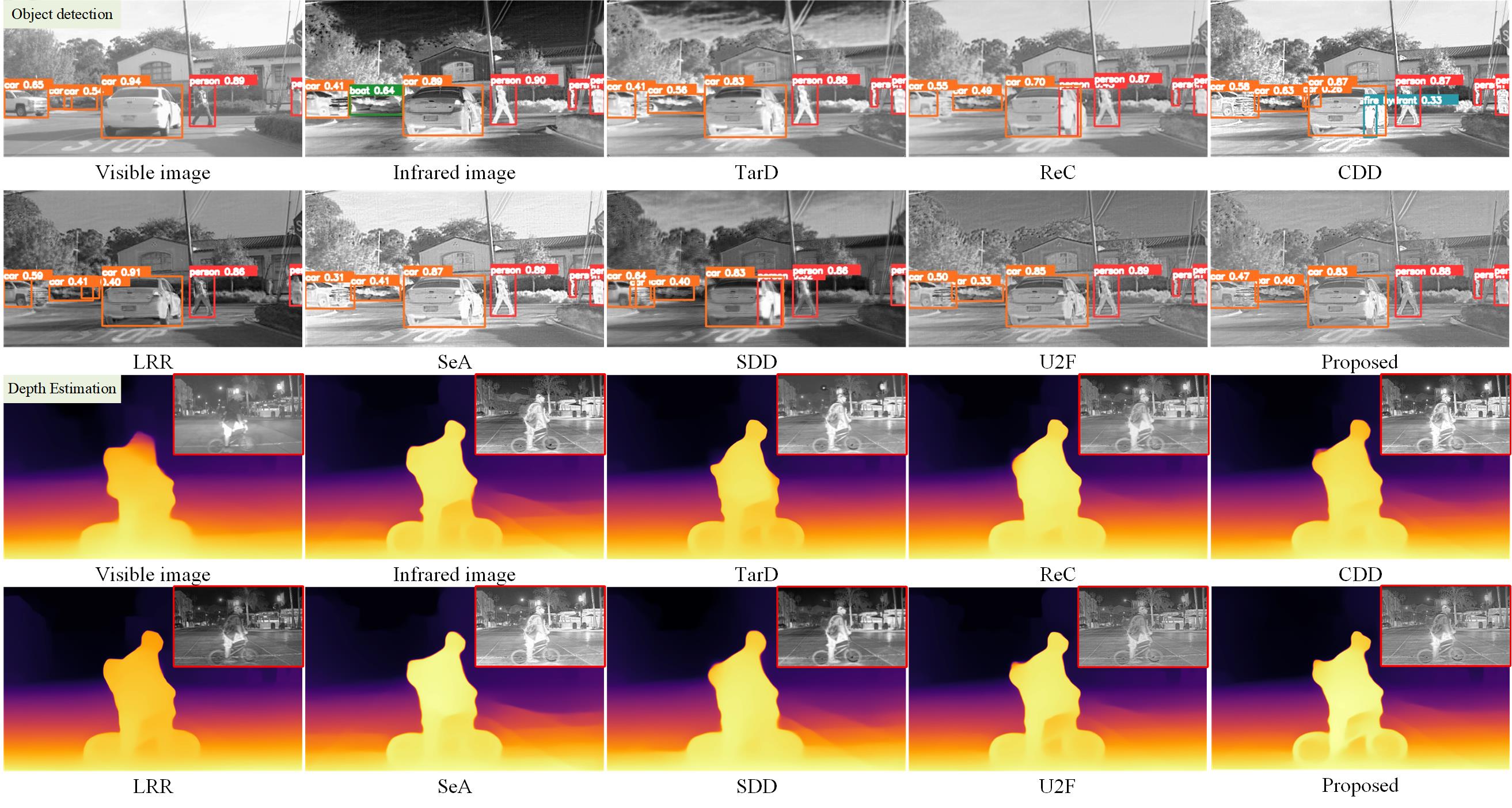}
   \caption{Visual comparison of different algorithms on other computer vision tasks.}
   \label{fig:onecol6}
\end{figure*}

\cref{tab:comparison2} presents the quantitative comparison results of all algorithms on the RoadScene dataset. The proposed algorithm ranked among the top two for four metrics, demonstrating superior fusion performance compared to the other algorithms. Notably, CDD, SeA and U2F also exhibited outstanding performance on this dataset, surpassing other methods in terms of detail retention and correlation with the source image.

\begin{table}[h]
\centering
\caption{Average scores by quantitative comparison on the RoadScene dataset. \textbf{Bold} is the best and \textcolor{red}{Red} is the second-best.}
\footnotesize 
\label{tab:comparison2}
\begin{tabular}{@{}l@{\hspace{1em}}llllll@{}}
\hline
Methods & $Q_G$ & $Q_M$ & $Q_S$ & $AG$ & $SF$ & $PSNR$ \\
\hline
TarD\cite{C50}(\textit{CVPR}) & 0.35 & 0.41 & 0.75 & 4.17 & 10.45 & 15.03 \\
ReC\cite{C46}(\textit{ECCV}) & 0.33 & 0.39 & 0.74 & 3.80 & 9.10 & 15.35 \\
CDD\cite{C47}(\textit{CVPR}) & 0.46 & 0.42 & 0.75 & \textbf{6.78} & \textbf{18.90} & 14.01 \\
LRR\cite{C43}(\textit{TPAMI}) & 0.32 & 0.36 & 0.59 & 4.64 & 12.37 & 11.81 \\
SeA\cite{C9}(\textit{Inf}) & 0.46 & \textcolor{red}{0.47} & 0.77 & \textcolor{red}{6.52} & \textcolor{red}{16.69} & 13.61 \\
SDD\cite{C25}(\textit{TMM}) & 0.37 & 0.39 & 0.74 & 4.38 & 10.44 & 12.89 \\
U2F\cite{C20}(\textit{TPAMI}) & \textcolor{red}{0.49} & 0.45 & \textbf{0.81} & 4.64 & 11.42 & \textbf{16.21} \\
\rowcolor{gray!20} Proposed & \textbf{0.50} & \textbf{0.57} & \textcolor{red}{0.80} & 4.97 & 13.57 & \textcolor{red}{16.17} \\
\hline
\end{tabular}
\end{table}

\subsection{Comparison results on the PET-MRI dataset}
We extend our experiments to multi-modal medical image fusion (MEIF) to investigate the generalization capability of the proposed algorithm. For MEIF, we can represent the model as a fusion process of visible images A and infrared images. We utilize the Harvard Medical School public database \footnote{\url{http://www.med.harvard.edu/aanlib/home.html}} as the dataset for this experiment, which contains 100 pairs of magnetic resonance imaging (MRI) and positron emission tomography (PET) images.

\cref{tab:comparison3} presents the quantitative comparison experiments between proposed algorithm and four state-of-the-art MEIF algorithms (LRD\cite{C33}, ReC\cite{C46}, MATR\cite{C35}, U2F\cite{C20}) on the PET-MRI fusion task. As shown in \cref{tab:comparison3}, proposed method outperforms the comparison method and achieves the highest score on all indexes. The results show that the proposed algorithm integrates useful information from different modal medical images and can be used to assist doctors in clinical diagnosis.

\begin{table}[h]
  \centering
  \caption{Quantitative comparison of five methods on PET-MRI fusion task. \textbf{Bold} is the best and \textcolor{red}{Red} is the second-best.}
  \footnotesize 
  \label{tab:comparison3}
  \begin{tabular}{@{}l@{\hspace{1em}}llllll@{}}
    \toprule
    Methods & $Q_G$ & $Q_M$ & $Q_S$ & $AG$ & $SF$ & $PSNR$ \\ \midrule
    LRD\cite{C33}(\textit{TIM})     & 0.25 & 0.08 & 0.25 & 9.04 & 27.63 & 8.32  \\
    MATR\cite{C35}(\textit{TIP})    & \textcolor{red}{0.64} & \textcolor{red}{0.27} & 0.54 & \textcolor{red}{9.13} & \textcolor{red}{28.04} & 13.07 \\
    ReC\cite{C46}(\textit{ECVV})    & 0.62 & 0.17 & \textcolor{red}{0.80} & 9.08 & 25.46 & \textcolor{red}{13.77} \\
    U2F\cite{C20}(\textit{TPAMI}) & 0.41 & 0.11 & 0.33 & 3.71 & 11.08 & 10.98 \\
    \rowcolor{gray!20} Proposed & \textbf{0.73} & \textbf{0.61} & \textbf{0.89} & \textbf{9.84} & \textbf{31.75} & \textbf{14.13} \\ \bottomrule
  \end{tabular}
\end{table}

\subsection{Applications in other computer vision tasks}
In this section, we assess the effectiveness of the proposed fusion method in other computer vision tasks: object detection and depth estimation. The target detection task is implemented using YoloV4 \cite{C31}, while the depth estimation task is based on MiDaS \cite{C32}. Among the algorithms evaluated, only SeA, U2F, and the proposed algorithm are able to accurately detect all significant target information within the scene. Other algorithms, such as ReC, CDD, and LRR, exhibit varying degrees of detection errors. \cref{fig:onecol6} illustrates that CDD, SeA, and SDD all fall short in providing improved depth information. In contrast, our proposed algorithm surpasses all compared algorithms by effectively leveraging advanced task-driven approaches. It provides comprehensive and accurate semantic and depth information for the scene, thereby greatly assisting in advanced vision tasks.

\section{Conclusion}
To solve the problems of incomplete image focus regions and difficulties in capturing salient information specific to different modalities simultaneously, a focused information integration framework for MMIF is proposed in this study. The proposed method first decomposes the source image into structural and textural components using the SSF. To fuse the texture layer, we employ a multi-scale feature detection guided by a pyramid, while energy information in the structural layer is extracted considering pixel distribution in multiple directions and using an effective fusion strategy based on entropy. Extensive experimental results show that the proposed method can yield the cutting-edge performance than current techniques.
This study not only effectively synthesises two different types of image fusion tasks, multi-focus and multi-modal, but also shows strong potential for other computer vision tasks include object detection and depth estimation.

\section{Acknowledgements}
This work was supported by the National Natural Science Foundation of China (Nos. 62201149, 52374166, and 62271148).
{\small
\bibliographystyle{ieee_fullname}
\bibliography{egbib}
}

\end{document}